\documentclass[lettersize,journal]{IEEEtran}
\usepackage{amsmath,amsfonts}
\usepackage{amssymb}
\usepackage{amsthm}
\theoremstyle{definition}
\newtheorem{definition}{Definition}

\newtheorem{remark}{Remark}
\usepackage{algorithmic}
\usepackage{algorithm}
\usepackage{array}
\usepackage[caption=false,font=footnotesize,labelfont=sf,textfont=sf]{subfig}
\usepackage{textcomp}
\usepackage{stfloats}
\usepackage{url}
\usepackage{verbatim}
\usepackage{graphicx}
\usepackage{cite}
\usepackage{mathtools}
\usepackage{siunitx}
\usepackage{booktabs}
\usepackage{multirow}
\usepackage{supertabular}
\usepackage{longtable}
\LTcapwidth=\textwidth

\usepackage{tikz}
\usetikzlibrary{matrix, backgrounds, fit}

\newcommand{\e}[1]{\emph{#1}}
\newcommand{\bftab}{\fontseries{b}\selectfont}

\newcommand{\mc}[1]{\mathcal{#1}}
\newcommand{\mb}[1]{\mathbb{#1}}

\newcommand{\lra}{\longrightarrow}
\newcommand{\lrma}{\longmapsto}
\newcommand{\missing}{\mathord{?}}

\usepackage{fancyhdr}
\begin{document}

\title{Polar Encoding: A Simple Baseline Approach for Classification with Missing Values}

\author{Oliver~Urs~Lenz,
        Daniel~Peralta,
        and~Chris~Cornelis
\thanks{The research reported in this paper was conducted with the financial support of the Odysseus programme of the Research Foundation -- Flanders (FWO).}%
\thanks{Oliver Urs Lenz is with the Leiden Institute of Advanced Computer Science, Leiden University, Leiden, 2333 CA, The Netherlands, and also with the Department of Applied Mathematics, Computer Science and Statistics, Ghent University, Ghent, 9000, Belgium (e-mail: o.u.lenz@liacs.leidenuniv.nl).}%
\thanks{Chris Cornelis is with the Department of Applied Mathematics, Computer Science and Statistics, Ghent University, Ghent, 9000, Belgium (e-mail: chris.cornelis@ugent.be).}
\thanks{Daniel Peralta is with the Department of Information Technology, Ghent University -- imec, Ghent, 9000, Belgium (e-mail: daniel.peralta@ugent.be).}}

\markboth{IEEE Transactions on Fuzzy Systems}%
{Shell \MakeLowercase{\textit{et al.}}: A Sample Article Using IEEEtran.cls for IEEE Journals}

\maketitle
\thispagestyle{fancy}

\begin{abstract}
We propose polar encoding, a representation of categorical and numerical $[0,1]$-valued attributes with missing values to be used in a classification context. We argue that this is a good baseline approach, because it can be used with any classification algorithm, preserves missingness information, is very simple to apply and offers good performance. In particular, unlike the existing missing-indicator approach, it does not require imputation, ensures that missing values are equidistant from non-missing values, and lets decision tree algorithms choose how to split missing values, thereby providing a practical realisation of the \e{missingness incorporated in attributes} (MIA) proposal. Furthermore, we show that categorical and $[0,1]$-valued attributes can be viewed as special cases of a single attribute type, corresponding to the classical concept of barycentric coordinates, and that this offers a natural interpretation of polar encoding as a fuzzified form of one-hot encoding. With an experiment based on twenty real-life datasets with missing values, we show that, in terms of the resulting classification performance, polar encoding performs better than the state-of-the-art strategies  \e{multiple imputation by chained equations} (MICE) and \e{multiple imputation with denoising autoencoders} (MIDAS) and --- depending on the classifier --- about as well or better than mean/mode imputation with missing-indicators.
\end{abstract}


\section{Introduction}
\label{sec_introduction}

\IEEEPARstart{M}{issing} values are a frequent issue in real-life datasets and a subject of ongoing research \cite{garcia20incremental,zhang22incomplete,li22hybrid}. In the present paper, we consider what a good baseline approach is for handling missing values in the context of classification.

Missing values have been extensively studied in the context of statistical inference. For estimating a parameter value, the generally accepted approach is to perform \e{multiple imputation} \cite{rubin78multiple}, in which one models the posterior distribution of the values that are missing on the basis of the non-missing values. By drawing from this distribution, one obtains a sample of imputed datasets and a corresponding sample of the estimand, allowing one to estimate the true parameter value and determine the uncertainty of this estimate due to the missing values. Two popular multiple imputation proposals are \e{multiple imputation by chained equations} (MICE) \cite{buuren99flexible} and \e{multiple imputation with denoising autoencoders} (MIDAS) \cite{lall22midas}.

The explicit assumption behind multiple imputation is that the distribution of missing values can be estimated on the basis of non-missing values (\e{missing at random} (MAR)). In contrast, the assumption behind the \e{missing-indicator} approach \cite{cohen68multiple} is that missingness is potentially informative (\e{missing not at random} (MNAR)), and that this aspect of the data should be explicitly represented through binary indicator attributes, that record for each original attribute whether the value was missing. If one assumes that missing values are not part of the `true' model, missing-indicators introduce bias \cite{jones96indicator}, and for this reason they have generally been dismissed in the context of statistical inference.

In the context of machine learning, and of classification in particular, model bias is arguably less important than prediction performance. We have previously established, through the first large-scale evaluation of missing-indicators on real-life datasets, that these do generally increase classification performance \cite{lenz22no}.

For decision trees, missingness is also preserved by the \e{missingness incorporated in attributes} (MIA) approach \cite{twala08good}, which stipulates that the tree construction algorithm should evaluate two versions of each split, with missing values included on either side. MIA has been shown to outperform imputation with or without missing indicators \cite{josse20consistency}.

We conclude from this that missing values are an important part of a dataset, that should be made available for classifiers to learn from just like non-missing values. While missing-indicators can be used for this, there are two aspects that prevent them from being an ideal baseline approach towards missing values. Both stem from the fact that missing-indicators have to be combined with imputation. Firstly, this means that the practitioner still needs to make a choice --- which imputation method to use. And secondly, while missing-indicators preserve missing values, we will see that to a certain extent, the imputation still induces the classifier to treat missing values like their imputed values.

To address this, we introduce in the present paper a new approach towards missing values called \e{polar encoding}, which can be used with categorical and $[0, 1]$-scaled numerical attributes. Polar encoding represents missing values without relying on imputation, leaving it completely up to the classifier how to learn from missing values. As we will see, polar encoding is a very simple proposal, but to the best of our knowledge, it has never been suggested before.

In theoretical terms, the motivation for our proposal can be understood through an analogy with Bayesian statistics. Choosing an imputation strategy encodes the prior knowledge of the practitioner about the value of the missing values. This can be useful when the practitioner has information about the mechanism responsible for the missing values. But in the absence of any such information, i.e. in the general case, a good baseline approach should let the practitioner avoid committing themself to any prior knowledge. Polar encoding enables that, just as so-called \e{noninformative} prior distributions do in Bayesian statistics, ``the rationale for [which] is often said to be `to let the data speak for [itself]''' \cite{gelman13bayesian}.

We will proceed by defining polar encoding, and comparing it against imputation and missing-indicators on the basis of four criteria that a good baseline approach towards missing values should satisfy (Section~\ref{sec_baseline}). Next, we specifically explain why polar encoding is a good approach for distance-based (Section~\ref{sec_generalisations}) and decision tree (Section~\ref{sec_mia}) classifiers. In Section~\ref{sec_barycentric}, we offer additional conceptual motivation for polar encoding by arguing that it can be seen as a fuzzification of one-hot encoding.

We complement these theoretical arguments in Section~\ref{sec_experimental_evaluation} with an experimental evaluation of the downstream classification performance of polar encoding, by comparing it against MICE and MIDAS, as well as against missing-indicators paired with mean/mode imputation, on the basis of twenty real-life datasets with missing values. Finally, we present our conclusions in Section~\ref{sec_conclusion}.

\section{Polar encoding as a good baseline approach}
\label{sec_baseline}

\begin{figure*}
\centering
\subfloat[Missing-indicator]{
\includegraphics[width=.31\linewidth]{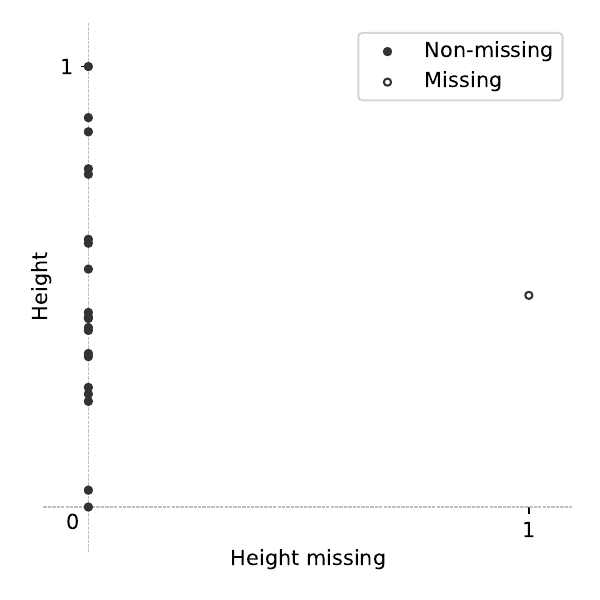}
\label{fig_missing_indicator}}%
\hfil
\subfloat[Boscovich polar encoding]{
\includegraphics[width=.31\linewidth]{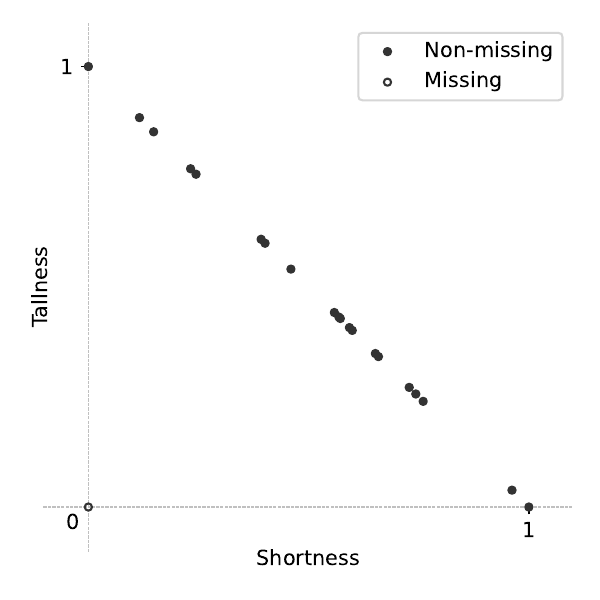}
\label{fig_fuzzy_one_hot}}%
\hfil
\subfloat[Euclidean polar encoding]{
\includegraphics[width=.31\linewidth]{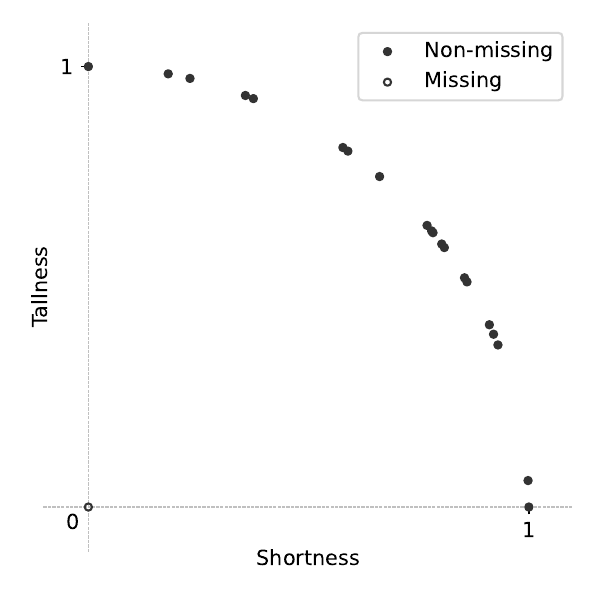}
\label{fig_euclidean_ohe}}%
\caption{Illustrative example of a $[0, 1]$-valued attribute for height with missing value, with missing-indicator and polar encoding.}
\label{fig_missing_value_encodings}
\end{figure*}

We will now present polar encoding\footnote{We have chosen the name \e{polar encoding} as a loose analogy to polar coordinates, because values are encoded in relation to a number of poles: the origin and the unit vectors $\left\langle 1, 0\right\rangle$ and $\left\langle 0, 1\right\rangle$ (and higher-dimensional unit vectors for categorical attributes).}, and discuss why it is a good baseline approach towards missing values. For comparison, one-hot encoding \cite{suits57use} is a simple baseline solution to the related problem of handling categorical attributes with algorithms that expect numerical input. It preserves the information encoded in categorical attributes and results in a dataset that can be fed to any numerical algorithm. Polar encoding is a similar solution, but for missing values.

For categorical attributes, polar encoding corresponds exactly to one-hot encoding, with missing values represented as zero vectors. Meanwhile, each $[0, 1]$-scaled numerical attribute is converted into a pair of features with the following map\footnote{This is the default form of polar encoding, to be used with (Boscovich) 1-distance and with algorithms not based on distance. We will propose a separate form of polar encoding to be used with (Euclidean) 2-distance in Subsection~\ref{sec_euclidean_distance}.}:

\begin{equation}
\label{eq_fohe_map}
\begin{aligned}
 x &\lrma \left\langle x, 1 - x\right\rangle,\\
 \missing &\lrma \left\langle 0, 0\right\rangle,
\end{aligned}
\end{equation}
where $x$ is any non-missing value, and $\missing$ a missing value. The resulting representation is illustrated by Fig.~\ref{fig_fuzzy_one_hot}, which contrasts with the representation produced by the missing-indicator approach (Fig.~\ref{fig_missing_indicator}).

We propose that in the context of classification, the qualities of a good baseline approach towards missing values are embodied by the following four criteria:

\textbf{Modularity.} The baseline approach should be self-contained. It should result in a complete, numerically encoded dataset, allowing classification algorithms to be agnostic about missing values.

\textbf{Conservatism.} The baseline approach should be a faithful representation of the original dataset. It should presuppose as little as possible about how missing values contribute to the learning task.

\textbf{Simplicity.} The baseline approach should be simple to apply. It should require a minimal amount of computational effort and no parameter choices by the user.

\textbf{Performance.} The baseline approach should enable good downstream prediction performance. It should perform well on average across real-life classification problems.

We may have to accept a certain trade-off between these criteria. For example, the simplicity and modularity of a good baseline approach may outweigh slightly lower downstream performance compared to a vastly more complicated solution. Conversely, we could accept a less conservative approach as a good baseline if it combined simplicity and modularity with superior performance. However, it turns out that in the context of supervised learning, conservatism, simplicity and performance appear to go somewhat hand in hand.

\begin{figure*}
\centering
\includegraphics[width=\linewidth]{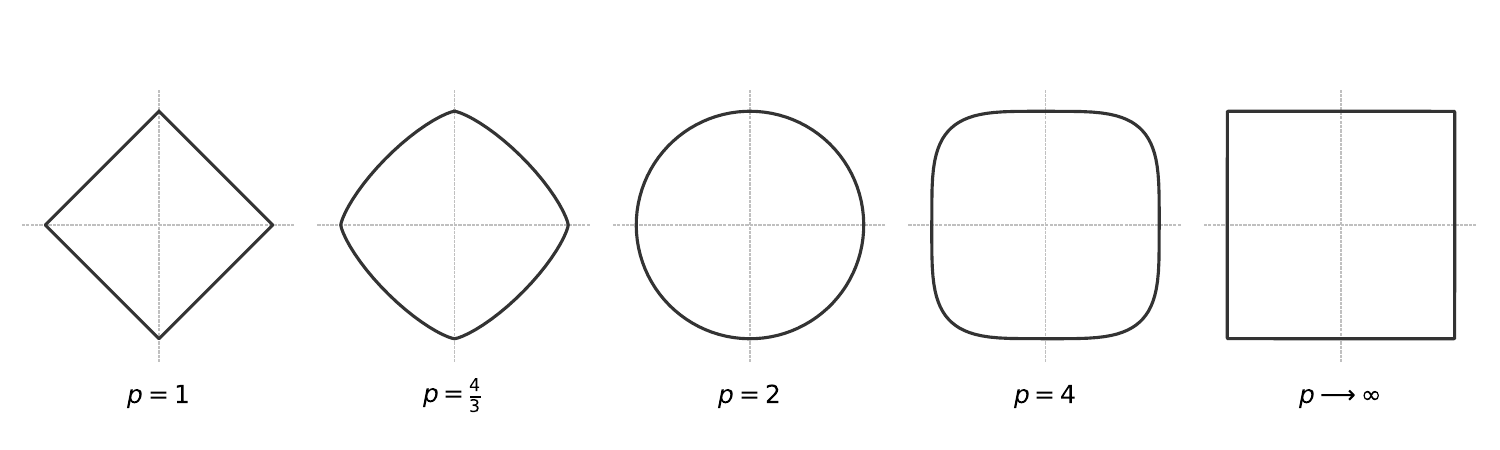}
\caption{Minkowski $p$-norm unit circles for various values of $p$.}
\label{fig_unit_spheres}
\end{figure*}

Imputation satisfies modularity, as it replaces missing values with estimates and the resulting complete dataset can be fed to any classification algorithm. However, by design, it is not a conservative approach towards missing values, since it predetermines the contribution of missing values towards the classification task by replacing them. Most imputation algorithms are not simple either, requiring substantial amounts of computation and user input.

Arguably the simplest form of imputation is imputation with the mean of numerical attributes and the mode of categorical attributes. Mean/mode imputation is not considered a good solution for statistical inference because it introduces bias. However, somewhat counterintuitively, it does not necessarily lead to worse prediction performance in the context of supervised learning \cite{perezlebel22benchmarking}. The reason for this is that missing values can be informative, and it is precisely because mean/mode imputation fails to hide missing values well that this information remains partially available for prediction algorithms to learn from.

The missing-indicator approach is more conservative than imputation, because it preserves missing values. It also generally increases classification performance on real-life datasets \cite{lenz22no}. However, the missing-indicator approach is not maximally conservative. As it has to be combined with imputation, it induces classification algorithms towards treating missing values like their imputed values. For algorithms that are based on distance, like nearest neighbours algorithms and support vector machines, the missing-indicator approach represents a missing value as being closer to its imputed value (e.g. the mean) than to other values (Fig.~\ref{fig_missing_indicator}). For decision tree algorithms, missing values will always split together with their imputed value when the algorithm splits on the original attribute.

Finally, because MIA lets decision trees choose how to split missing values, it is a conservative approach. However, it is not modular, since it requires an adaptation of the prediction algorithm itself, and because it can only be used with decision trees.

In contrast to these existing proposals, polar encoding satisfies all four criteria. Polar encoding is modular, since it results in a complete, numerical dataset that can be used with any classification algorithm. It is also simple. Being essentially a linear transformation of the data, it can be applied quickly and easily, without the need for any dedicated software. In the next two sections, we will show that polar encoding is conservative. In particular, we will argue that by making missing values equidistant from non-missing values, polar encoding does not presuppose their contribution to the classification problem, and that for decision trees, it is effectively a modular implementation of MIA. Finally, we will show in Section~\ref{sec_experimental_evaluation} that the performance of polar encoding is also as good or better than the alternatives.

\section{Polar encoding and distance-based classifiers}
\label{sec_generalisations}

In this section, we will explain how polar encoding ensures that missing values are equidistant from all non-missing values, and present a variant proposal for Euclidean distance.

\subsection{Boscovich distance}
\label{sec_boscovich_distance}

Recall the general definition of the Minkowski $p$-norm of a vector $x \in \mb{R}^m$, for $p \geq 1$:

\begin{equation}
 \left|x\right|_p := \left(\sum_{i \leq m} \left\vert x_i \right \vert^p\right)^{\frac{1}{p}}.
\end{equation}

The Minkowski $p$-distance between any two points $x, y \in \mb{R}^m$ is the $p$-norm of their difference. The $p$-norm unit sphere in $\mb{R}^m$ consists of all points with $p$-norm equal to 1. For $m = 2$, this gives us the $p$-norm unit circles (Fig.~\ref{fig_unit_spheres}). 

Two values of $p$ are particularly often used in machine learning. When $p = 1$, we obtain the Boscovich norm\footnote{Perhaps first used implicitly by Roger Joseph Boscovich (1711--1787) to minimise regression residuals \cite{boscovich57litteraria,boscovich60recentissimis,todhunter73history,eisenhart61boscovich,stigler86history}; also known as \e{city block}, \e{Manhattan}, \e{rectilinear} and \e{taxicab} norm.}, which reduces to $\sum_{i \leq m} \left\vert x_i \right \vert$, and when $p = 2$, we obtain the Euclidean norm\footnote{Also known as Pythagorean norm.}.

Fig.~\ref{fig_fuzzy_one_hot} illustrates the application of polar encoding with a toy example. The key observation to make is that unlike the missing-indicator approach, the Boscovich distance between a missing value and any non-missing value is always 1. In fact, this is a simple consequence of the fact that polar encoding maps non-missing values onto the non-negative quadrant of the Boscovich unit circle.

Moreover, with polar encoding, the Boscovich distance between any two non-missing values $x, y \in [0, 1]$ becomes twice the original distance $\left\vert x - y\right\rvert$. In other words, the distances between non-missing values remain essentially unchanged, except for a scaling factor of 2. The Boscovich distance between a missing value and non-missing values is 1, which is exactly half the maximum distance 2 between two non-missing values, reflecting the fact that we do not know what the `true' value of a missing value is. This distance can be used directly, or transformed into a similarity value with $a \lrma 1 - a/2$. In this case, the similarity between a missing value and any non-missing value is always 0.5, exactly half the maximum similarity of 1.

This contrasts with the approach taken in \cite{dai13rough}, where the similarity between a missing value and any other value is stipulated to always be 1. Similarly, the authors of the present paper have proposed \cite{lenz21adapting} (based on previous work \cite{jensen09interval}) to propagate the uncertainty from missing values using interval-valued fuzzy sets. These interval values are bounded by an optimistic scenario, corresponding to the proposal in \cite{dai13rough}, and a pessimistic scenario, in which the similarity between a missing value and any other value (possibly also missing) is 0 (complete dissimilarity). In both cases the problem is that missing values are not more similar to each other than to non-missing values --- missing values are not treated as a signal to generalise from. Moreover, in practice these similarity relations scale poorly to larger datasets, because they do not admit straightforward implementations in terms of an existing distance measure.

\subsection{Euclidean distance}
\label{sec_euclidean_distance}

Based on the discussion in the previous subsection, a straightforward way to obtain polar encoding for Euclidean distance is to map non-missing values onto the non-negative quadrant of the Euclidean unit circle (Fig.~\ref{fig_euclidean_ohe}). We propose to do this with the following mapping, which establishes a linear correspondence between distance in $[0, 1]$ and arc length (scaling by a factor $\sqrt{2}$): 

\begin{equation}
\label{eq_fohe_2_map}
\begin{aligned}
 x &\lrma \left\langle\sin\frac{x\cdot \pi}{2}, \cos\frac{x\cdot \pi}{2}\right\rangle,\\
 \missing &\lrma \left\langle 0, 0\right\rangle.
\end{aligned}
\end{equation}

Note that this map cannot preserve Euclidean distance. When encoding a $[0, 1]$-valued attribute in this manner, larger distances become relatively less large. However, the difference is relatively small and may not be problematic in practice. For instance, the Euclidean distance between the minimum and maximum values becomes $\sqrt{2} \approx 1.41$, which is slightly less than twice the distance between either value and the midrange:

\begin{equation*}
\left(\left\lvert \sin\frac{\pi}{4} - 0 \right\rvert^2 + \left\lvert \cos\frac{\pi}{4} - 1 \right\rvert^2\right)^{\frac{1}{2}} \approx 0.765.\\
\end{equation*}

Furthermore, this maximum distance between two non-missing values ($\sqrt{2}$), is now comparatively smaller than with Boscovich distance (2). This is completely consistent with the distance between two different one-hot encoded categorical values, which is likewise $\sqrt{2}$ for Euclidean distance and 2 for Boscovich distance.

For other values of $p$, there exist generalisations of $\sin$ and $\cos$ that could be used instead to parametrise the non-negative quadrant of the $p$-unit sphere \cite{shelupsky59generalization,lindqvist00parclength}. However, these functions are defined as the inverses of integrals, and so are not easy to apply in practice.

\section{Polar encoding and decision tree classifiers}
\label{sec_mia}

\begin{figure}
\centering
\includegraphics[width=\linewidth]{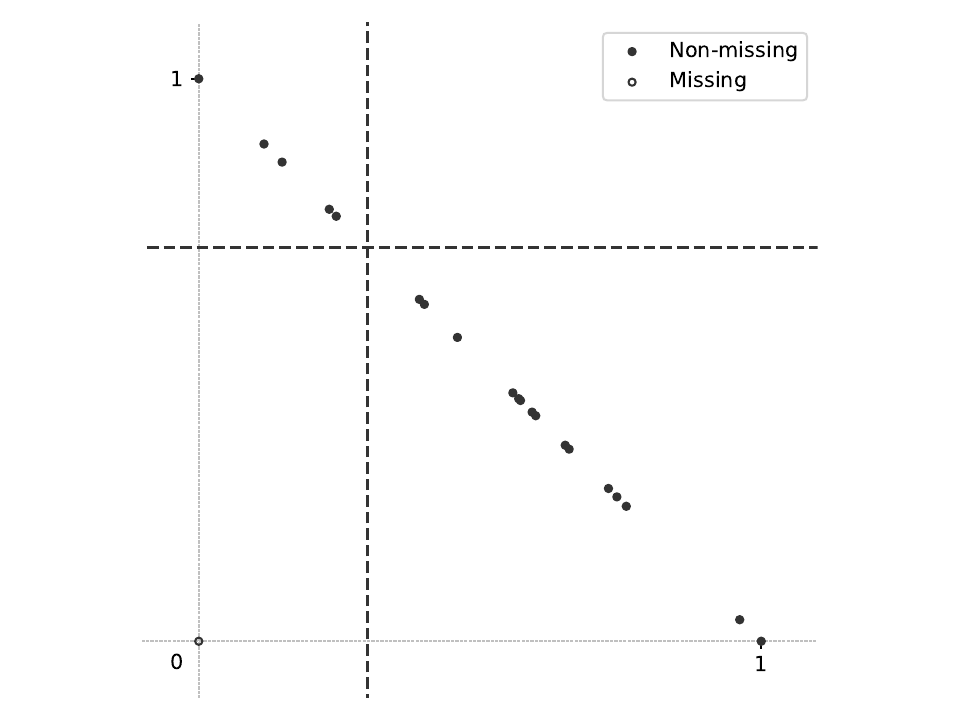}
\caption{Illustrative example of equivalent splits on a polar-encoded attribute, with missing values on either side.}
\label{fig_polar_splits}
\end{figure}

Polar encoding also allows decision tree algorithms to learn from missing values. The two dimensions of a polar-encoded attribute induce identical splits on the data, except that missing values end up on either side of each split (Fig.~\ref{fig_polar_splits}). Therefore, decision trees are effectively offered a choice as to which side of each split missing values should be grouped with. Missing values can also be split off on their own by splitting on both dimensions of a polar-encoded attribute.

This contrasts with the missing-indicator approach, where missing values either group together with their imputed value (when the tree splits on the original attribute), or alone (when the tree splits on the missing-indicator).

The effect of polar encoding on decision trees is very similar to the \e{missingness incorporated in attributes} (MIA) proposal \cite{twala08good} which stipulates that when splitting on an attribute with missing values, the algorithm should consider each potential split twice, with missing values on either side, and additionally a split that separates non-missing and missing values. MIA has been added to the scikit-learn \cite{pedregosa11scikitlearn} implementation of LightGBM \cite{ke17lightgbm}, and a similar strategy is part of XGBoost \cite{chen16xgboost}. The advantage of polar encoding is that it can be applied by the user, and combined with off-the-shelf implementations of decision tree algorithms that do not natively support MIA.\footnote{A similar trick is suggested in \cite{josse20consistency}: repeat each attribute with missing features twice, and encode missing values alternatively as $-\infty$ and $+\infty$.}

The performance of MIA has mostly been evaluated on the basis of simulated data with informative missing values.

For decision trees, MIA performs better \cite{twala08good} than resolving missing values as a weighted combination of the two branches according to the prior probabilities of the non-missing values \cite{cestnik87assistant}, and about as good as multiple imputation with expectation maximisation \cite{schafer97analysis}, which had emerged as the two best-performing strategies in a previous comparison \cite{twala09empirical}.

For Bayesian additive regression trees, MIA has been shown to outperform random forest imputation \cite{kapelner15prediction}. Similarly, MIA has been shown to outperform mean imputation with missing-indicators and a handful of other strategies for regression with decision trees, Random Forest and XGBoost \cite{josse20consistency}.

Finally the scikit-learn implementation of LightGBM mentioned above has also been evaluated on four large, real-life medical datasets, and MIA was found to produce somewhat to considerably better performance than the missing-indicator approach with various forms of imputation \cite{perezlebel22benchmarking}.

\section{Polar encoding as representation of barycentric attributes}
\label{sec_barycentric}

\begin{figure}
\centering
\subfloat[Crisp partition]{
\makebox[.4\linewidth][c]{
$\left(\begin{array}{@{}c|c|c@{}}
0 & 0 & 1\\
0 & 0 & 1\\
1 & 0 & 0\\
0 & 1 & 0\\
1 & 0 & 0\\
0 & 0 & 1\\
\end{array}\right)$
}
\label{fig_crisp_partition}}%
\hfil
\subfloat[Categorical attribute]{
\makebox[.4\linewidth][c]{
$\left(\begin{array}{@{}ccc@{}}
0 & 0 & 1\\\hline
0 & 0 & 1\\\hline
1 & 0 & 0\\\hline
0 & 1 & 0\\\hline
1 & 0 & 0\\\hline
0 & 0 & 1\\
\end{array}\right)$
}
\label{fig_crisp_attribute}}%
\caption{Example illustrating the correspondence between crisp partitions and categorical attributes of a dataset. Rows correspond to the records, columns to the partition classes and categories. The values 1 and 0 indicate membership and non-membership, respectively.}
\label{fig_crisp_tables}
\end{figure}

In this section, we will show how polar encoding can be seen as the representation of \e{barycentric} attributes, which generalise both categorical and $[0, 1]$-valued attributes. In particular, this explains how polar encoding generalises one-hot encoding. To begin with, we establish our working definitions of datasets, attributes and one-hot encoding.

\subsection{Numerical and categorical attributes}
\label{sec_one_hot_encoding}

A key difference between numerical and categorical attributes is that while the values of numerical attributes can be assumed to lie in $\mb{R}$, allowing us to construct machine learning models based on the arithmetic of $\mb{R}$, the set of values $V$ of a categorical attribute is not assumed to have any relevant internal structure.

However, many algorithms are only defined for numerical data, and one popular solution, perhaps first documented by Suits~(1957) \cite{suits57use} (but ``not new'' even then), is to transform a categorical attribute into a tuple of numerical features through \e{one-hot} encoding (or encoding with \e{dummy variables}).

\begin{definition}
\label{def_ohe}
Let $V$ be a categorical attribute. For a chosen order $V = (v_1, v_2, \dots, v_p)$, its \e{(redundant) one-hot encoding} is the map $V \lra [0, 1]^p$ that sends $v_i$ to the standard basis vector $\mathbf{e}_i = \left\langle 0,  \dots, 0, 1, 0, \dots, 0 \right\rangle$ for all $i \leq p$, while its \e{compact one-hot encoding} is the map $V \lra [0, 1]^{p - 1}$ that sends $v_p$ to $\mathbf{0}$ and $v_i$ for $i < p$ to $\mathbf{e}_i$.
\end{definition}

Compact one-hot encoding is sufficient to ensure that all categorical values are linearly separable, but it also introduces an asymmetry that can be undesirable.

\begin{remark}
\label{rem_binary}
Binary attributes can be represented both as categorical attributes and as numerical attributes. In the latter case, a typical choice is to use the values 0 and 1. This numerical representation corresponds directly to a compact one-hot encoding of its categorical representation. We will exploit this correspondence to argue that barycentric attributes generalise not just categorical, but also $[0, 1]$-valued numerical attributes.
\end{remark}

It is a classical observation that categorical attributes correspond to partitions \cite{quinlan86induction}. Formally, a categorical attribute $V$ induces a partition on a dataset $X$ through the equivalence relation that equates elements of $X$ with the same value in $V$. Conversely, if we have a partition $\mc{U}$ of $X$, we can derive a categorical attribute of $X$ that takes, for each $x \in X$, the value $U$ in $\mc{U}$ that contains $y$.

Both categorical attributes (through one-hot encoding) and partitions can be represented with a matrix of values in $\{0, 1\}$, with exactly one value equal to 1 on each row (Fig.~\ref{fig_crisp_tables}). In Subsection~\ref{sec_fuzzy_categorical_attributes}, we will extend this correspondence between categorical attributes and partitions to barycentric attributes and fuzzy partitions.

\subsection{Barycentric attributes}
\label{sec_barycentric_attributes}

\begin{figure}
\centering
\includegraphics[width=\linewidth]{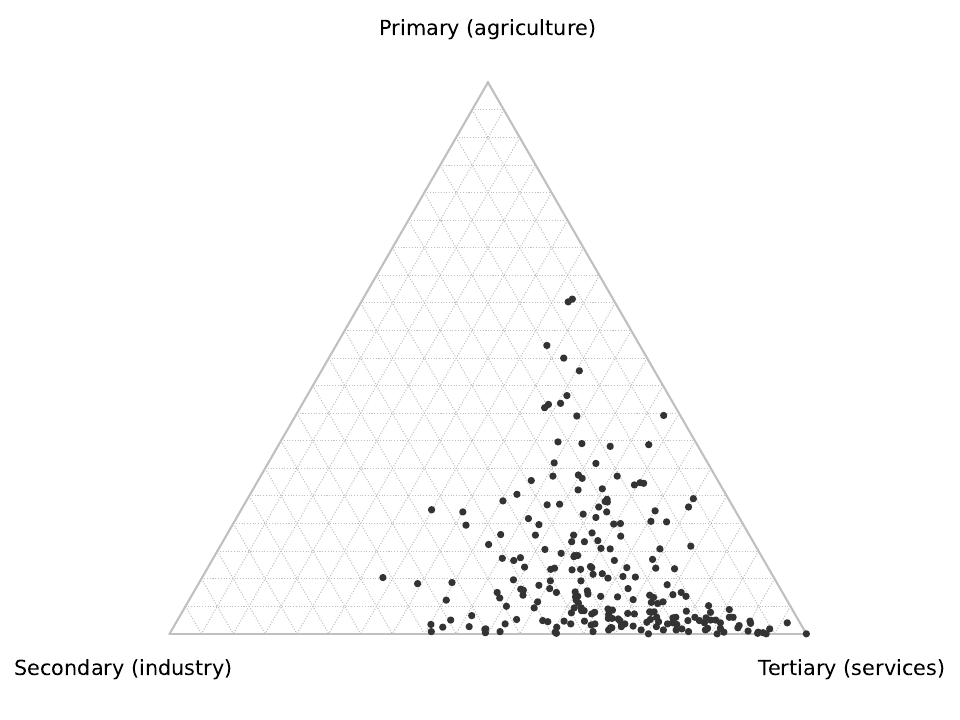}
\caption{Example of a ternary plot: distribution of GDP over economic sectors of countries and territories \cite{cia22gdp}.}
\label{fig_ternary_plot}
\end{figure}

Barycentric values (or \e{coordinates}; also known as \e{homogeneous} coordinates) are numerical values that sum to a fixed number (typically 1), or where only the relative proportions are considered important. The concept dates back to at least Möbius~(1827) \cite{mobius27barycentrische,allardice91barycentric,boyer56golden}, who used it to express a point as the weighted sum (the \e{barycentre}, where the weights cancel each other out) of the vertices of a simplex. Barycentric values are also used to define the points that make up projective space. For the purpose of the present paper, we will assume that barycentric values are non-negative, and use the following formal definition:

\begin{definition}
An attribute is \e{barycentric} if it is equal to a copy of $\left(\mb{R}_{\geq 0}^m \setminus \{\mathbf{0}\}\right) / \sim$ for some $m \geq 1$, where $\sim$ is the equivalence relation defined by $\left\langle x_1, x_2, \dots, x_m \right\rangle \sim \left\langle \lambda x_1, \lambda x_2, \dots, \lambda x_m \right\rangle$ for all $\lambda \in \mb{R}_{>0}$. The \e{normalised representation} of a value $[x_1, x_2, \dots, x_m] \in \left(\mb{R}_{\geq 0}^m \setminus \{\mathbf{0}\}\right) / \sim$ is the vector $\left\langle x_1/s, x_2/s, \dots, x_m/s \right\rangle \in \mb{R}^m$, where $s := \sum_{i \leq m} x_i$.
\end{definition}

Barycentric values are often encountered in the literature in the form of ternary plots (Fig.~\ref{fig_ternary_plot}), which display the relative frequencies of three components. Recent examples include the composition of planets (core, mantle and hydrosphere) \cite{huang22magrathea,macdonald22confirming,haldemann23exoplanet}, seabed sediment \cite{wang21study}, ternary mixtures of fluids \cite{stemplinger21theory,tonsmann21surface}, ternary compounds \cite{chen21machine,nolan21computationguided} and even human behaviour \cite{kim21win,molter22gazedependent}.

In addition, some machine learning problems are typically approached by considering relative token frequencies. For instance, this can be part of the calculation of the cosine similarity between text records \cite{zhao18fuzzy,tian21adaptive,sangma23fhcnds}.

Finally, the confidence scores produced by a classification model (or some other estimate), when normalised to sum to 1, are also a natural example of barycentric values.

\subsection{Barycentric attributes as fuzzified categorical attributes}
\label{sec_fuzzy_categorical_attributes}

\begin{figure}
\centering
\subfloat[Fuzzy partition]{
\makebox[.4\linewidth][c]{
$\left(\begin{array}{@{}c|c|c@{}}
0.2 & 0.6 & 0.2\\
0.1 & 0.9 & 0.0\\
0.4 & 0.1 & 0.5\\
1.0 & 0.0 & 0.0\\
0.1 & 0.9 & 0.0\\
0.2 & 0.2 & 0.6\\
\end{array}\right)$
}
\label{fig_fuzzy_partition}}%
\hfil
\subfloat[Barycentric attribute]{
\makebox[.4\linewidth][c]{
$\left(\begin{array}{@{}ccc@{}}
0.2 & 0.6 & 0.2\\\hline
0.1 & 0.9 & 0.0\\\hline
0.4 & 0.1 & 0.5\\\hline
1.0 & 0.0 & 0.0\\\hline
0.1 & 0.9 & 0.0\\\hline
0.2 & 0.2 & 0.6\\
\end{array}\right)$
}
\label{fig_fuzzy_attribute}}%
\caption{Example illustrating the correspondence between fuzzy partitions and barycentric attributes of a dataset. Rows correspond to the records, columns to the partition classes and categories. Values are membership degrees.}
\label{fig_fuzzy_tables}
\end{figure}

Barycentric attributes generalise categorical attributes in the following way. If $\left(\mb{R}_{\geq 0}^m \setminus \{\mathbf{0}\}\right) / \sim$ is a barycentric attribute, then the subset $V$ of values with only one non-zero coefficient forms a categorical attribute, and we will write $B(V) := \left(\mb{R}_{\geq 0}^m \setminus \{\mathbf{0}\}\right) / \sim$ and say that $V$ is the set of categories of $B(V)$. In particular, the normalised representation of $B(V)$ reduces precisely to one-hot encoding when restricted to $V$.

This relationship can also be understood geometrically. The set of normalised representations of a barycentric attribute coincides with the standard $m-1$-simplex, which is spanned by $m$ vertices, the one-hot encoded values of $V$.

Conversely, barycentric attributes can be understood as fuzzified categorical attributes, allowing us to give a fuzzy answer to the question of category membership:

\begin{remark}
\label{ref_fca}
Let $B(V)$ be a barycentric attribute with $m$ categories. Then we can associate to each value in $B(V)$ with normal representation $\left\langle x_1, x_2, \dots, x_m \right\rangle$ the fuzzy set in $V$ with membership degrees $x_1, x_2, \dots, x_m$. These are precisely the fuzzy sets in $V$ with cardinality 1.
\end{remark}

This is reinforced by the fact that barycentric attributes correspond to fuzzy partitions in the same way that categorical attributes correspond to crisp partitions (Subsection~\ref{sec_one_hot_encoding}). Recall the definition of a fuzzy partition \cite{ruspini69new,dunn74fuzzy}:

\begin{definition}
Let $X$ be a finite set. A \e{fuzzy partition} on $X$ is a finite set $\mc{F}$ of fuzzy sets in $X$ such that for each $x \in X$, we have $\sum_{F \in \mc{F}} F(x) = 1$.
\end{definition}

To see that a barycentric attribute $B(V)$ on a dataset $X$ contains the same information as a fuzzy partition on $X$, consider that both can be represented by a $|X| \times |V|$ matrix of values in $[0, 1]$, such that the rows sum to 1 \cite{bezdek78fuzzy}. The columns of such a matrix correspond to a fuzzy partition (Fig.~\ref{fig_fuzzy_partition}), whereas its rows correspond to the normalised values of a barycentric attribute (Fig.~\ref{fig_fuzzy_attribute}).

\subsection{$[0, 1]$-valued attributes as barycentric attributes}
\label{sec_generalisation_numerical_attributes}

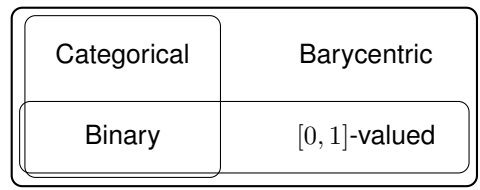
\begin{figure}
\centering
\begin{tikzpicture}[mytext/.style={text=black, font=\sffamily\normalsize}]

\matrix (N) [ 
    matrix of nodes,
    nodes={mytext, anchor=center, rounded corners},     
    column sep=10mm,
    row sep= 5mm,]{
        \makebox[2cm]{Categorical} & \makebox[2cm]{Barycentric} \\
        \makebox[2cm]{Binary} & \makebox[2cm]{$[0, 1]$-valued} \\
    };

\begin{scope}[on background layer]

\node[fit=(N-1-1) (N-2-1), inner ysep=7pt, inner xsep=5pt, rounded corners, draw] (aux) {};

\node[fit=(N-1-1) (N-2-1) (N-1-2) (N-2-2), inner xsep=10pt, inner ysep=10pt, thick, draw, rounded corners] {};

\node[fit=(N-2-1) (N-2-2), inner ysep=5pt, inner xsep=7pt, draw, rounded corners] {};

\end{scope}
\end{tikzpicture}
\caption{Euler diagram of different attribute types. Barycentric attributes generalise both categorical and $[0, 1]$-valued attributes.}
\label{fig_attributes}
\end{figure}

Just as one-hot encoding is redundant and we can use compact one-hot encoding to represent the same information with one fewer value (Definition~\ref{def_ohe}), so the normalised representation of a barycentric attribute $\left\langle x_1, x_2, \dots, x_m \right\rangle$ is redundant, and we can encode it compactly as $\left\langle x_1, x_2, \dots, x_{m-1} \right\rangle$. Together, these compactly encoded values form the $m-1$-simplex in $\mb{R}^{m-1}$ spanned by the standard $m-2$-simplex and the origin. Conversely, we can reconstruct the full representation from a compactly encoded value $\left\langle x_1, x_2,\allowbreak \dots,\allowbreak x_{p-1} \right\rangle$ by appending the value $1 - \sum_{i \leq p - 1} x_i$. 

The compact encoding of a barycentric attribute with only two categories is a single value in $[0, 1]$. This leads us to the following observation:

\begin{remark}
\label{rem_fuzzy_binary}
 Let $A$ be a $[0, 1]$-valued attribute. Then the values of $A$ are compactly encoded values of a barycentric attribute with two categories. We obtain the corresponding redundant representation with $x \lrma \left\langle x, 1 - x \right\rangle$. Thus, barycentric attributes generalise not just categorical attributes, but also $[0, 1]$-valued attributes (Fig.~\ref{fig_attributes}).
\end{remark}

This redundant representation of $[0, 1]$-valued attributes generalises the categorical representation of binary attributes that we noted in Remark~\ref{rem_binary}. We can illustrate this with an example. Suppose that we have a binary attribute denoting height, with two values, `short' and `tall'. Its compact encoding is as a single numerical attribute $A$ with two values, 0 and 1, expressing `tallness'. Its redundant encoding is as two numerical attributes, tallness ($A$) and shortness ($1 - A$). Likewise, suppose that we have $[0, 1]$-valued attribute $A'$ denoting height, then its redundant encoding $\left\langle A', 1 - A' \right\rangle$ consists of fuzzy expressions of `tallness' and `shortness'.

Of course, this redundant encoding of a $[0, 1]$-valued attribute is precisely the polar encoding that we propose in this paper (Fig.~\ref{fig_fuzzy_one_hot}).

\subsection{Representing missing values}
\label{sec_missing_values}

We now turn to the representation of missing values. Recall our example from the previous subsection: suppose that we have a barycentric attribute $B(V)$ denoting height, with $V$ containing the two categories `tall' and `short', then a missing value does not convey positive information about either category. Therefore, we accommodate the possibility that a barycentric attribute can have a missing value by expanding the set $\left(\mb{R}_{\geq 0}^m \setminus \{\mathbf{0}\}\right) / \sim$ to $\mb{R}_{\geq 0}^m / \sim$, and by stipulating that the normalised representation of $[0, 0, \dots, 0]$ is the zero vector $\mathbf{0}$. This corresponds to the unique fuzzy set in $V$ with cardinality 0 (the empty set).

Note that barycentric attributes with missing values can no longer be represented compactly, since doing so would also encode the non-missing value $[0, 0, \dots, 1]$ as $\mathbf{0}$. It is precisely the redundancy of the redundant normal representation (in particular, redundant one-hot encoding) that enables us to encode missing values as zeroes. For $[0, 1]$-valued numerical attributes, this means that our proposed polar encoding is necessary if we want to represent missing values.

\begin{table}
\centering
\caption{Classification Algorithms}
\label{tab_classifiers}
\begin{tabular}{p{.105\linewidth}p{.805\linewidth}}
\toprule
\multicolumn{2}{l}{Distance-based classifiers} \\
\midrule
     NN & Nearest Neighbours \cite{fix51discriminatory}\\
     NN-D & Nearest Neighbours, distance-weighted \cite{dudani76distance}\\
     FRNN & Fuzzy Rough Nearest Neighbours \cite{jensen08new} with OWA \cite{cornelis10ordered}\\
     SVM-G & Soft-margin Support Vector Machine \cite{cortes95support} with Gaussian kernel \\
\midrule
\multicolumn{2}{l}{Decision tree classifiers} \\
\midrule
      CART & Classification and Regression Tree \cite{breiman84classification}\\
        RF & Random Forest\cite{breiman01random}\\
       ERT & Extremely Randomised Trees\cite{geurts06extremely}\\
       ABT & Ada-Boosted Trees \cite{freund95desiciontheoretic} with SAMME \cite{zhu09multiclass}\\
       GBM & Gradient Boosting Machine \cite{friedman01greedy}\\
\bottomrule
\end{tabular}
\end{table}

\section{Experimental evaluation}
\label{sec_experimental_evaluation}

We now describe our experimental evaluation of using polar encoding for classification. Concretely, we ask whether it leads to better classification performance than the sophisticated imputation strategies MICE and MIDAS and mean/mode imputation with missing-indicators.

\subsection{Setup}
\label{sec_experimental_setup}

For MICE, we will use the recent \e{miceforest} implementation for Python \cite{wilson20miceforest}, which employs LightGBM \cite{ke17lightgbm} to obtain predictions, while for MIDAS, we use the \e{MIDASpy} implementation for Python \cite{lall23efficient}. Since we want to obtain a single dataset that can be used as input for various classification algorithms, we perform single rather than multiple imputation. Otherwise, we use default hyperparameter values. We use our own implementations of mean/mode imputation with missing-indicators and polar encoding, in the latter case by manually applying the transformations \eqref{eq_fohe_map} and \eqref{eq_fohe_2_map} in Python.

\begin{table}
\centering
\caption{Real-life datasets with missing values (adapted from \cite{lenz22no}).}
\label{tab_statistics}
\begin{tabular}{p{.27\linewidth}rrll}
\toprule
                   Dataset & Records &  Attributes & Missing rate &                          Source \\
\midrule
              \mbox{adult} & 48\,842 &          13 &        0.010 &          \cite{kohavi96scaling} \\
   \mbox{agaricus-lepiota} &    8124 &          22 &        0.014 &       \cite{schlimmer87concept} \\
        \mbox{aps-failure} & 76\,000 &         170 &        0.083 &       \cite{ferreiracosta16ida} \\
         \mbox{arrhythmia} &     443 &         279 &       0.0032 &      \cite{guvenir97supervised} \\
              \mbox{bands} &     540 &          34 &        0.054 &        \cite{evans94overcoming} \\
                \mbox{ckd} &     400 &          24 &         0.11 &       \cite{rubini15generating} \\
                \mbox{crx} &     690 &          15 &       0.0065 &     \cite{quinlan87simplifying} \\
        \mbox{dress-sales} &     500 &          12 &         0.19 &                                 \\
            \mbox{exasens} &     399 &           7 &         0.43 &   \cite{soltanizarrin20invitro} \\
                \mbox{hcc} &     165 &          49 &         0.10 &              \cite{santos15new} \\
      \mbox{heart-disease} &    1611 &          14 &         0.17 &   \cite{detrano89international} \\
          \mbox{hepatitis} &     155 &          19 &        0.057 &       \cite{efron81statistical} \\
        \mbox{horse-colic} &     368 &          20 &         0.26 &       \cite{mcleish90enhancing} \\
\mbox{mammographic-masses} &     961 &           4 &        0.042 &        \cite{elter07prediction} \\
                 \mbox{mi} &    1700 &         111 &        0.085 & \cite{golovenkin20trajectories} \\
              \mbox{nomao} & 34\,465 &         118 &         0.38 &       \cite{candillier12design} \\
      \mbox{primary-tumor} &     330 &          17 &        0.039 &       \cite{cestnik87assistant} \\
              \mbox{secom} &    1567 &         590 &        0.045 &        \cite{mccann08causality} \\
            \mbox{soybean} &     683 &          35 &        0.098 &      \cite{michalski80learning} \\
        \mbox{thyroid0387} &    9172 &          23 &        0.069 &       \cite{quinlan86inductive} \\
\bottomrule
\end{tabular}
\end{table}

We evaluate our selection of missing data approaches for two sets of classifiers: distance-based and decision tree--based algorithms (Table~\ref{tab_classifiers}). For the Support Vector Machine with Gaussian kernel that is based on Euclidean distance, we evaluate the Euclidean variant of polar encoding, while for the nearest neighbour algorithms that allow setting the distance measure as a hyperparameter, we evaluate both the standard and the Euclidean variant.

We use the same collection of twenty datasets from the UCI repository for machine learning \cite{dua19uci} with naturally occurring missing values that we previously used in \cite{lenz22no} (Table~\ref{tab_statistics}). These datasets show great variation --- they cover a number of different domains and contain between 155 and 76\,000 records, between 4 and 590 attributes, between 2 and 21 decision classes and missing value rates between 0.0032 and 0.43. We rescale numerical attributes to $[0, 1]$, before applying polar encoding or imputation. In the latter case, we then also apply one-hot encoding to categorical attributes.

We evaluate classification performance using the area under the receiver operating characteristic (AUROC) \cite{hand01simple}. For each dataset, we perform five-fold stratified cross-validation, repeat this five times for different random divisions of the data, and take the mean of the resulting 25 AUROC scores. To establish whether the performance of polar encoding vis-\`a-vis imputation generalises to other (similar) datasets, we test for significance using one-sided Wilcoxon signed-ranks tests \cite{wilcoxon45individual}. A $p$-value below 0.5 indicates that polar encoding performed better, while a $p$-value above 0.5 indicates that it performed worse.

For all classifiers we use the implementations provided by the Python library \e{scikit-learn} \cite{pedregosa11scikitlearn}, except for FRNN, where we use our own implementation in \e{fuzzy-rough-learn} \cite{lenz22fuzzyroughlearn}. For our main experiment, we use default hyperparameter values, with three exceptions informed by the findings in \cite{lenz22no}: with CART we perform cost complexity pruning ($\alpha = 0.01$), with ERT we set the number of trees to 1000, and with GBM we apply early-stopping.

We also perform a follow-up experiment in which we compare polar encoding against mean/mode imputation with missing-indicator for the same set of classifiers but with hyperparameter optimisation. For NN, NN-D and FRNN, we optimise $k$ for all values in the range $[1, 50]$. For SVM, we optimise $C$ and $\gamma$ by randomly drawing 10 pairs of values from the exponential distribution $\frac{1}{\beta}e^{-\frac{1}{\beta}x}$, with, respectively, scale $\beta = 100$ and scale $\beta = \frac{1}{10}$. For the decision tree classifiers, we optimise the number of features that are considered at each split, as well as the minimum number of records required to continue splitting nodes, by randomly drawing 10 pairs of values from the interval $[0, 1]$, interpreted as share of the total number of features or records. For NN, NN-D and FRNN, we apply efficient leave-one-out validation, whereas for the other classifiers we apply stratified (nested) five-fold cross-validation,\footnote{For datasets with classes that contain only four records in the training set, we have applied four-fold cross-validation instead.} selecting the hyperparameter values that result in the highest (mean) validation AUROC.

\subsection{Results}
\label{sec_results}

Table~\ref{tab_p_values} lists the $p$-values obtained from comparing the performance of polar encoding against the performance of MICE, MIDAS and mean/mode imputation with missing-indicators, in terms of the mean AUROC for each classifier and each dataset.\footnote{The mean AUROC scores are provided as supplementary material.}

\begin{table}
\centering
\caption{$p$-values, polar encoding vs other missing value approaches.}
\label{tab_p_values}
\begin{tabular}{llllll}
\toprule
Distance & Classifier & \multicolumn{4}{l}{Alternative}\\
\cmidrule{3-6}
 & & MICE & MIDAS & \multicolumn{2}{l}{\vtop{\hbox{\strut Mean/mode imputation}\hbox{\strut with missing-indicators}}}\\
\cmidrule{3-6}
 &  & \multicolumn{4}{l}{Hyperparameter values}\\
\cmidrule{3-6}
 & & Default & Default & Default & Optimised\\
\midrule
Boscovich & NN &      0.011 &  0.024 &      0.18 &  0.074 \\
    & NN-D &      0.011 &  0.068 &      0.19 &   0.16 \\
    & FRNN &     0.0088 &  0.049 &    0.0024 & 0.0019 \\
\midrule
Euclidean & NN &     0.0098 & 0.0070 &      0.14 &   0.19 \\
    & NN-D &      0.018 &  0.024 &      0.15 &  0.086 \\
    & FRNN &     0.0056 & 0.0056 &    0.0040 & 0.0021 \\
    & SVM-G &     0.0027 & 0.0035 &     0.018 &  0.039 \\
\midrule
--- & CART &       0.13 &  0.085 &     0.031 &  0.058 \\
    & RF &      0.012 &   0.23 &      0.40 &   0.57 \\
    & ERT &     0.0063 &   0.20 &      0.14 &   0.23 \\
    & ABT &     0.0045 &  0.054 &     0.054 &   0.77 \\
    & GBM &     0.0017 &  0.099 &      0.61 &   0.50 \\
\bottomrule
\end{tabular}
\end{table}

The first thing to note is that with default hyperparameter values and for our selection of datasets, polar encoding generally increases classification performance, except for RF and GBM, where it leads to approximately the same performance as mean/mode imputation with missing-indicators. On the whole, the $p$-values for Euclidean distance are not higher than the $p$-values for Boscovich distance, which indicates that the relative advantage of polar encoding is not less with Euclidean distance. In Subsection~\ref{sec_euclidean_distance}, we noted that the Euclidean variant of polar encoding introduces a slight distortion to the distances between non-missing values, but this does not appear to be harmful for classification performance.

\begin{table}
\centering
\caption{$p$-values, polar encoding vs other missing value approaches.}
\label{tab_total_p_values}
\begin{tabular}{lllll}
\toprule
\vtop{\hbox{\strut Distance used with}\hbox{\strut NN, NN-D and FRNN}} & \multicolumn{3}{l}{Alternative}\\
\cmidrule{2-5}
 & MICE & MIDAS & \multicolumn{2}{l}{\vtop{\hbox{\strut Mean/mode imputation}\hbox{\strut with missing-indicators}}}\\
\cmidrule{2-5}
 & \multicolumn{4}{l}{Hyperparameter values}\\
\cmidrule{2-5}
 & Default & Default & Default & Optimised\\
\midrule
Boscovich &     0.0012 &  0.011 &     0.011 & 0.047 \\
Euclidean &     0.0012 & 0.0043 &     0.011 & 0.044 \\
\bottomrule
\end{tabular}
\end{table}

Not all of the $p$-values are significant, which may be due to the small sample size (20 datasets). Overall, the advantage of polar encoding over pure imputation is more pronounced than the advantage of polar encoding over mean/mode imputation with missing-indicators, which agrees with our previous finding that missing-indicators increase performance because they preserve missingness-information. Nevertheless, it appears that the greater conservatism of polar encoding gives classifiers even more opportunity to learn from missing values. If we perform a clustered Wilcoxon signed-rank test \cite{rosner06wilcoxon} on the scores obtained for all datasets and all classifiers, clustered by dataset, we find that polar encoding performs significantly better than the alternative approaches, regardless of whether we use Boscovich or Euclidean distance (Table~\ref{tab_total_p_values}).

Table~\ref{tab_p_values} also contains the $p$-values from our follow-up experiment, testing the performance of polar encoding against mean/mode imputation with missing-indicators in the context of hyperparameter optimisation. The $p$-values are essentially similar to the $p$-values obtained with default hyperparameter optimisation, except for ABT, where mean/mode imputation with missing-indicators now has a slight advantage. Across all classifiers, polar encoding still performs better (Table~\ref{tab_total_p_values}).

\section{Conclusion}
\label{sec_conclusion}

In this paper we have presented polar encoding, a novel method to represent missing values of categorical and $[0, 1]$-valued attributes. We have argued that in the context of classification, it presents a good baseline approach for missing values because it is modular, conservative, simple and performant.

In particular, polar encoding is more conservative than the current baseline, mean/mode imputation with missing-indicators, because it does not just preserve the information from missing values, but also does not pre-suppose their contribution to the classification problem, as it avoids imputation altogether. For distance-based algorithms, it ensures that missing values are equidistant from all non-missing values. For decision tree algorithms, it allows missing values to be grouped on either side of each split. This latter behaviour corresponds to the existing MIA approach, with the crucial difference that polar encoding can be combined with all sorts of classification algorithms, which do not have to be adapted for this purpose.

We have provided further justification for polar encoding by showing that it can be viewed as a fuzzification of one-hot encoding, the standard approach for representing categorical attributes numerically. We did this by formalising the concept of barycentric attributes, which can be seen as both a fuzzification of categorical attributes and a generalisation of $[0, 1]$-valued attributes. Because one-hot encoding is slightly redundant, using one more dimension than strictly necessary, it allows us to represent missing values as zero vectors, symbolising the absence of information.

Having previously shown that missing-indicators improve classification performance on real-life datasets, in the present paper we conducted an experiment to test whether polar encoding works even better. We found that in the context of classification, polar encoding generally outperforms two sophisticated imputation algorithms, MICE and MIDAS. Polar encoding also performs better than mean/mode imputation with missing indicators, although this difference is less pronounced, and mean/mode imputation may have a slight advantage with ABT if hyperparameter optimisation is applied.

In the future, we would like to extend polar encoding to numerical attributes that are scaled differently. For example, when an attribute is scaled by its standard deviation, polar encoding could be adapted to ensure that missing values are equidistant from all `typical' non-missing values, namely those contained within one standard deviation of the mean.

\bibliographystyle{IEEEtran}
\bibliography{IEEEabrv,20240215_polar_encoding}

\clearpage
\onecolumn
\appendix[Full results]

We list here the mean AUROC across five-fold cross-validation and five random states for each classifier, each dataset, and each missing value approach, for distance-based classifiers and decision tree classifiers with default hyperparameter values (Tables~\ref{tab_nn_auroc} and \ref{tab_dt_auroc}, respectively) and with optimised hyperparameter values (Tables~\ref{tab_nn_optimised_auroc} and \ref{tab_dt_optimised_auroc}, respectively). \textbf{MMI-I}: mean/mode imputation with missing indicators. \textbf{PE}: polar encoding.

\begin{longtable}{llp{.063\linewidth}p{.063\linewidth}p{.063\linewidth}p{.063\linewidth}p{.063\linewidth}p{.063\linewidth}p{.063\linewidth}p{.063\linewidth}}
\caption{Distance-based classifiers, default hyperparameter values. \textbf{Bold}: highest value per distance measure.}
\label{tab_nn_auroc}\\
\toprule
Classifier & Dataset & \multicolumn{4}{l}{Boscovich distance} & \multicolumn{4}{l}{Euclidean distance} \\
      &  &    MICE &         MIDAS &     MMI-I &         PE &    MICE &         MIDAS &     MMI-I &         PE \\
\midrule
\endfirsthead
\caption[]{Distance-based classifiers, default hyperparameter values. \textbf{Bold}: highest value per distance measure.} \\
\toprule
Classifier & Dataset & \multicolumn{4}{l}{Boscovich distance} & \multicolumn{4}{l}{Euclidean distance} \\
      &  &    MICE &         MIDAS &     MMI-I &         PE &    MICE &         MIDAS &     MMI-I &         PE \\
\midrule
\endhead
\midrule
\multicolumn{10}{r}{{Continued on next page}} \\
\midrule
\endfoot

\bottomrule
\endlastfoot
NN & adult &         0.846 &         0.846 &         0.846 &  \bftab 0.846 &         0.846 &         0.846 &         0.846 &  \bftab 0.846 \\
      & agaricus-lepiota &  \bftab 1.000 &  \bftab 1.000 &  \bftab 1.000 &  \bftab 1.000 &  \bftab 1.000 &  \bftab 1.000 &  \bftab 1.000 &  \bftab 1.000 \\
      & aps-failure &         0.908 &  \bftab 0.910 &         0.910 &         0.909 &         0.895 &         0.897 &         0.902 &  \bftab 0.904 \\
      & arrhythmia &         0.756 &  \bftab 0.758 &         0.757 &         0.757 &         0.720 &         0.706 &  \bftab 0.733 &         0.723 \\
      & bands &         0.778 &         0.777 &         0.800 &  \bftab 0.824 &         0.770 &         0.766 &         0.794 &  \bftab 0.810 \\
      & ckd &         0.994 &         0.995 &         0.996 &  \bftab 0.999 &         0.997 &  \bftab 0.999 &         0.994 &         0.997 \\
      & crx &         0.909 &         0.909 &         0.910 &  \bftab 0.912 &         0.909 &         0.911 &         0.910 &  \bftab 0.911 \\
      & dress-sales &         0.535 &         0.545 &  \bftab 0.560 &         0.552 &         0.536 &         0.545 &  \bftab 0.560 &         0.548 \\
      & exasens &         0.697 &         0.712 &         0.717 &  \bftab 0.719 &         0.702 &  \bftab 0.719 &         0.713 &         0.717 \\
      & hcc &         0.718 &         0.703 &  \bftab 0.751 &         0.717 &         0.696 &         0.687 &  \bftab 0.733 &         0.699 \\
      & heart-disease &         0.833 &         0.836 &         0.833 &  \bftab 0.841 &         0.824 &         0.825 &         0.827 &  \bftab 0.832 \\
      & hepatitis &  \bftab 0.834 &         0.826 &         0.815 &         0.818 &  \bftab 0.839 &         0.827 &         0.815 &         0.828 \\
      & horse-colic &         0.735 &  \bftab 0.754 &         0.723 &         0.728 &  \bftab 0.735 &         0.732 &         0.727 &         0.730 \\
      & mammographic-masses &         0.820 &         0.822 &  \bftab 0.831 &         0.830 &         0.820 &         0.821 &  \bftab 0.830 &         0.830 \\
      & mi &         0.551 &         0.559 &  \bftab 0.591 &         0.575 &         0.553 &         0.557 &  \bftab 0.584 &         0.583 \\
      & nomao &         0.960 &         0.970 &         0.980 &  \bftab 0.982 &         0.954 &         0.964 &         0.978 &  \bftab 0.980 \\
      & primary-tumor &         0.692 &         0.687 &  \bftab 0.719 &         0.687 &         0.695 &         0.687 &  \bftab 0.718 &         0.697 \\
      & secom &  \bftab 0.623 &         0.622 &         0.590 &         0.617 &         0.591 &  \bftab 0.598 &         0.522 &         0.548 \\
      & soybean &         0.976 &         0.988 &         0.988 &  \bftab 0.992 &         0.973 &         0.986 &         0.987 &  \bftab 0.990 \\
      & thyroid0387 &         0.797 &         0.819 &         0.833 &  \bftab 0.835 &         0.785 &         0.804 &         0.827 &  \bftab 0.830 \\
\midrule
NN-D & adult &         0.825 &         0.825 &         0.826 &  \bftab 0.828 &         0.827 &         0.826 &  \bftab 0.827 &         0.827 \\
      & agaricus-lepiota &  \bftab 1.000 &  \bftab 1.000 &  \bftab 1.000 &  \bftab 1.000 &  \bftab 1.000 &  \bftab 1.000 &  \bftab 1.000 &  \bftab 1.000 \\
      & aps-failure &         0.909 &         0.911 &  \bftab 0.911 &         0.910 &         0.896 &         0.897 &         0.903 &  \bftab 0.905 \\
      & arrhythmia &         0.759 &  \bftab 0.760 &         0.759 &         0.760 &         0.722 &         0.708 &  \bftab 0.735 &         0.726 \\
      & bands &         0.803 &         0.800 &         0.824 &  \bftab 0.851 &         0.784 &         0.780 &         0.808 &  \bftab 0.825 \\
      & ckd &         0.994 &         0.995 &         0.997 &  \bftab 0.999 &         0.997 &  \bftab 0.999 &         0.996 &         0.997 \\
      & crx &         0.905 &         0.905 &         0.906 &  \bftab 0.909 &         0.906 &         0.906 &         0.906 &  \bftab 0.909 \\
      & dress-sales &         0.534 &         0.543 &  \bftab 0.564 &         0.552 &         0.535 &         0.539 &  \bftab 0.563 &         0.548 \\
      & exasens &         0.694 &  \bftab 0.702 &         0.636 &         0.637 &         0.699 &  \bftab 0.708 &         0.632 &         0.634 \\
      & hcc &         0.738 &         0.723 &  \bftab 0.762 &         0.738 &         0.713 &         0.705 &  \bftab 0.744 &         0.720 \\
      & heart-disease &         0.835 &         0.839 &         0.837 &  \bftab 0.843 &         0.828 &         0.829 &         0.832 &  \bftab 0.837 \\
      & hepatitis &  \bftab 0.828 &         0.825 &         0.823 &         0.821 &  \bftab 0.833 &         0.829 &         0.819 &         0.827 \\
      & horse-colic &         0.750 &  \bftab 0.770 &         0.747 &         0.754 &         0.745 &  \bftab 0.750 &         0.745 &         0.749 \\
      & mammographic-masses &         0.791 &         0.801 &         0.808 &  \bftab 0.808 &         0.791 &         0.801 &         0.808 &  \bftab 0.808 \\
      & mi &         0.552 &         0.559 &  \bftab 0.592 &         0.577 &         0.554 &         0.558 &  \bftab 0.586 &         0.585 \\
      & nomao &         0.961 &         0.971 &         0.981 &  \bftab 0.983 &         0.954 &         0.965 &         0.979 &  \bftab 0.981 \\
      & primary-tumor &         0.676 &         0.681 &  \bftab 0.703 &         0.679 &         0.676 &         0.680 &  \bftab 0.704 &         0.688 \\
      & secom &  \bftab 0.630 &         0.628 &         0.594 &         0.624 &         0.594 &  \bftab 0.599 &         0.526 &         0.549 \\
      & soybean &         0.976 &         0.988 &         0.988 &  \bftab 0.992 &         0.974 &         0.986 &         0.987 &  \bftab 0.990 \\
      & thyroid0387 &         0.798 &         0.820 &         0.835 &  \bftab 0.837 &         0.787 &         0.805 &         0.829 &  \bftab 0.832 \\
\midrule
FRNN & adult &         0.872 &         0.871 &         0.872 &  \bftab 0.878 &         0.862 &         0.862 &         0.863 &  \bftab 0.867 \\
      & agaricus-lepiota &  \bftab 1.000 &  \bftab 1.000 &  \bftab 1.000 &  \bftab 1.000 &  \bftab 1.000 &  \bftab 1.000 &  \bftab 1.000 &  \bftab 1.000 \\
      & aps-failure &  \bftab 0.980 &         0.965 &         0.943 &         0.952 &  \bftab 0.975 &         0.964 &         0.962 &         0.968 \\
      & arrhythmia &         0.882 &         0.882 &  \bftab 0.889 &         0.887 &         0.856 &         0.855 &         0.868 &  \bftab 0.875 \\
      & bands &         0.812 &         0.810 &         0.832 &  \bftab 0.852 &         0.796 &         0.795 &         0.819 &  \bftab 0.833 \\
      & ckd &         1.000 &         1.000 &         0.999 &  \bftab 1.000 &  \bftab 1.000 &         1.000 &         0.998 &         1.000 \\
      & crx &         0.914 &         0.914 &         0.918 &  \bftab 0.921 &         0.914 &         0.915 &         0.918 &  \bftab 0.920 \\
      & dress-sales &         0.562 &         0.583 &  \bftab 0.592 &         0.577 &         0.558 &         0.566 &  \bftab 0.586 &         0.572 \\
      & exasens &         0.727 &         0.740 &         0.719 &  \bftab 0.745 &         0.727 &         0.740 &         0.736 &  \bftab 0.749 \\
      & hcc &         0.775 &         0.778 &         0.784 &  \bftab 0.792 &         0.777 &         0.771 &         0.769 &  \bftab 0.780 \\
      & heart-disease &         0.858 &         0.858 &         0.858 &  \bftab 0.863 &         0.849 &         0.846 &         0.848 &  \bftab 0.854 \\
      & hepatitis &  \bftab 0.887 &         0.884 &         0.882 &         0.884 &  \bftab 0.883 &         0.878 &         0.879 &         0.880 \\
      & horse-colic &         0.759 &  \bftab 0.794 &         0.760 &         0.772 &         0.761 &  \bftab 0.792 &         0.766 &         0.772 \\
      & mammographic-masses &         0.800 &         0.813 &         0.816 &  \bftab 0.838 &         0.806 &         0.816 &         0.824 &  \bftab 0.837 \\
      & mi &         0.668 &         0.680 &         0.674 &  \bftab 0.687 &         0.658 &         0.662 &         0.670 &  \bftab 0.678 \\
      & nomao &         0.976 &         0.983 &         0.986 &  \bftab 0.990 &         0.971 &         0.978 &         0.987 &  \bftab 0.989 \\
      & primary-tumor &  \bftab 0.794 &         0.787 &         0.794 &         0.790 &         0.788 &         0.780 &  \bftab 0.791 &         0.784 \\
      & secom &  \bftab 0.693 &         0.689 &         0.642 &         0.673 &         0.629 &  \bftab 0.630 &         0.596 &         0.609 \\
      & soybean &         0.992 &         0.997 &  \bftab 0.997 &         0.997 &         0.991 &         0.996 &         0.997 &  \bftab 0.997 \\
      & thyroid0387 &         0.871 &         0.872 &         0.888 &  \bftab 0.908 &         0.875 &         0.875 &         0.891 &  \bftab 0.902 \\
\midrule
SVM-G & adult &               &               &               &               &         0.892 &         0.892 &         0.893 &  \bftab 0.900 \\
      & agaricus-lepiota &               &               &               &               &  \bftab 1.000 &  \bftab 1.000 &  \bftab 1.000 &  \bftab 1.000 \\
      & aps-failure &               &               &               &               &         0.957 &         0.957 &         0.942 &  \bftab 0.974 \\
      & arrhythmia &               &               &               &               &         0.866 &         0.869 &         0.872 &  \bftab 0.878 \\
      & bands &               &               &               &               &         0.810 &         0.808 &         0.833 &  \bftab 0.843 \\
      & ckd &               &               &               &               &  \bftab 1.000 &         1.000 &         0.999 &         1.000 \\
      & crx &               &               &               &               &         0.918 &         0.920 &         0.920 &  \bftab 0.922 \\
      & dress-sales &               &               &               &               &         0.623 &         0.623 &  \bftab 0.632 &         0.614 \\
      & exasens &               &               &               &               &         0.760 &         0.767 &         0.768 &  \bftab 0.774 \\
      & hcc &               &               &               &               &         0.784 &         0.785 &  \bftab 0.800 &         0.789 \\
      & heart-disease &               &               &               &               &         0.860 &         0.862 &         0.861 &  \bftab 0.869 \\
      & hepatitis &               &               &               &               &  \bftab 0.861 &         0.853 &         0.857 &         0.858 \\
      & horse-colic &               &               &               &               &         0.759 &         0.786 &         0.776 &  \bftab 0.788 \\
      & mammographic-masses &               &               &               &               &         0.831 &         0.837 &  \bftab 0.845 &         0.835 \\
      & mi &               &               &               &               &         0.642 &         0.638 &         0.648 &  \bftab 0.655 \\
      & nomao &               &               &               &               &         0.981 &         0.986 &         0.990 &  \bftab 0.991 \\
      & primary-tumor &               &               &               &               &  \bftab 0.792 &         0.782 &         0.781 &         0.789 \\
      & secom &               &               &               &               &         0.702 &  \bftab 0.703 &         0.678 &         0.696 \\
      & soybean &               &               &               &               &         0.997 &         0.999 &         0.999 &  \bftab 0.999 \\
      & thyroid0387 &               &               &               &               &         0.877 &         0.874 &         0.894 &  \bftab 0.922 \\
\end{longtable}

\begin{longtable}{llp{.063\linewidth}p{.063\linewidth}p{.063\linewidth}p{.063\linewidth}}
\caption{Decision tree classifiers, default hyperparameter values. \textbf{Bold}: highest value.}
\label{tab_dt_auroc}\\
\toprule
Classifier & Dataset &    MICE &         MIDAS &     MMI-I &         PE \\
\midrule
\endfirsthead
\caption[]{Decision tree classifiers, default hyperparameter values. \textbf{Bold}: highest value.} \\
\toprule
Classifier & Dataset &    MICE &         MIDAS &     MMI-I &         PE \\
\midrule
\endhead
\midrule
\multicolumn{6}{r}{{Continued on next page}} \\
\midrule
\endfoot

\bottomrule
\endlastfoot
ABT & adult &         0.915 &         0.915 &         0.915 &  \bftab 0.915 \\
   & agaricus-lepiota &  \bftab 1.000 &  \bftab 1.000 &  \bftab 1.000 &  \bftab 1.000 \\
   & aps-failure &         0.986 &         0.986 &  \bftab 0.987 &         0.987 \\
   & arrhythmia &         0.633 &         0.634 &         0.634 &  \bftab 0.634 \\
   & bands &         0.805 &         0.806 &         0.806 &  \bftab 0.813 \\
   & ckd &         0.998 &         0.999 &         1.000 &  \bftab 1.000 \\
   & crx &         0.904 &         0.906 &         0.906 &  \bftab 0.908 \\
   & dress-sales &         0.585 &  \bftab 0.596 &         0.581 &         0.583 \\
   & exasens &         0.700 &         0.712 &         0.720 &  \bftab 0.722 \\
   & hcc &         0.713 &         0.722 &         0.725 &  \bftab 0.729 \\
   & heart-disease &         0.849 &         0.854 &  \bftab 0.860 &         0.859 \\
   & hepatitis &         0.798 &         0.777 &         0.807 &  \bftab 0.809 \\
   & horse-colic &         0.740 &  \bftab 0.763 &         0.756 &         0.756 \\
   & mammographic-masses &         0.852 &         0.855 &  \bftab 0.857 &         0.856 \\
   & mi &         0.576 &         0.568 &         0.572 &  \bftab 0.582 \\
   & nomao &         0.979 &         0.985 &         0.987 &  \bftab 0.987 \\
   & primary-tumor &  \bftab 0.662 &         0.662 &         0.660 &         0.648 \\
   & secom &         0.673 &  \bftab 0.673 &         0.668 &         0.663 \\
   & soybean &         0.744 &         0.851 &         0.870 &  \bftab 0.892 \\
   & thyroid0387 &         0.656 &         0.684 &         0.685 &  \bftab 0.685 \\
\midrule
CART & adult &  \bftab 0.844 &  \bftab 0.844 &  \bftab 0.844 &  \bftab 0.844 \\
   & agaricus-lepiota &         0.991 &  \bftab 0.993 &         0.992 &         0.992 \\
   & aps-failure &         0.860 &  \bftab 0.867 &         0.859 &         0.859 \\
   & arrhythmia &         0.746 &         0.747 &  \bftab 0.748 &         0.745 \\
   & bands &         0.742 &         0.731 &         0.759 &  \bftab 0.768 \\
   & ckd &         0.980 &  \bftab 0.980 &         0.975 &         0.977 \\
   & crx &  \bftab 0.900 &         0.897 &         0.896 &         0.897 \\
   & dress-sales &         0.572 &         0.566 &         0.570 &  \bftab 0.574 \\
   & exasens &         0.721 &         0.715 &         0.732 &  \bftab 0.743 \\
   & hcc &  \bftab 0.617 &         0.587 &         0.588 &         0.590 \\
   & heart-disease &         0.778 &  \bftab 0.780 &         0.777 &         0.774 \\
   & hepatitis &         0.645 &  \bftab 0.661 &         0.578 &         0.596 \\
   & horse-colic &         0.702 &         0.699 &  \bftab 0.723 &         0.718 \\
   & mammographic-masses &         0.815 &         0.816 &  \bftab 0.823 &         0.822 \\
   & mi &         0.535 &         0.581 &         0.592 &  \bftab 0.607 \\
   & nomao &         0.917 &  \bftab 0.935 &         0.916 &         0.916 \\
   & primary-tumor &         0.707 &         0.700 &         0.707 &  \bftab 0.739 \\
   & secom &  \bftab 0.500 &  \bftab 0.500 &  \bftab 0.500 &  \bftab 0.500 \\
   & soybean &         0.959 &         0.984 &         0.991 &  \bftab 0.993 \\
   & thyroid0387 &         0.877 &         0.884 &  \bftab 0.909 &         0.908 \\
\midrule
ERT & adult &         0.847 &         0.847 &         0.847 &  \bftab 0.856 \\
   & agaricus-lepiota &  \bftab 1.000 &  \bftab 1.000 &  \bftab 1.000 &  \bftab 1.000 \\
   & aps-failure &         0.991 &         0.991 &         0.991 &  \bftab 0.991 \\
   & arrhythmia &         0.897 &         0.899 &  \bftab 0.899 &         0.899 \\
   & bands &         0.879 &         0.888 &         0.890 &  \bftab 0.904 \\
   & ckd &         1.000 &         1.000 &  \bftab 1.000 &  \bftab 1.000 \\
   & crx &         0.915 &         0.916 &         0.914 &  \bftab 0.916 \\
   & dress-sales &         0.572 &         0.579 &  \bftab 0.602 &         0.575 \\
   & exasens &         0.716 &  \bftab 0.740 &         0.626 &         0.627 \\
   & hcc &         0.778 &         0.791 &  \bftab 0.808 &         0.803 \\
   & heart-disease &         0.857 &  \bftab 0.864 &         0.862 &         0.861 \\
   & hepatitis &         0.874 &  \bftab 0.879 &         0.873 &         0.857 \\
   & horse-colic &         0.776 &  \bftab 0.803 &         0.782 &         0.796 \\
   & mammographic-masses &         0.789 &         0.793 &         0.802 &  \bftab 0.805 \\
   & mi &         0.680 &         0.690 &         0.695 &  \bftab 0.709 \\
   & nomao &         0.985 &         0.990 &         0.994 &  \bftab 0.994 \\
   & primary-tumor &         0.698 &  \bftab 0.727 &         0.714 &         0.712 \\
   & secom &         0.745 &         0.740 &         0.746 &  \bftab 0.747 \\
   & soybean &         0.997 &         0.999 &         0.999 &  \bftab 0.999 \\
   & thyroid0387 &         0.975 &         0.983 &         0.988 &  \bftab 0.991 \\
\midrule
GBM & adult &         0.927 &         0.927 &  \bftab 0.927 &         0.927 \\
   & agaricus-lepiota &  \bftab 1.000 &  \bftab 1.000 &  \bftab 1.000 &  \bftab 1.000 \\
   & aps-failure &         0.988 &         0.987 &  \bftab 0.988 &         0.987 \\
   & arrhythmia &         0.848 &         0.848 &  \bftab 0.852 &         0.851 \\
   & bands &         0.846 &         0.859 &         0.857 &  \bftab 0.859 \\
   & ckd &         0.994 &         0.991 &         0.996 &  \bftab 0.996 \\
   & crx &  \bftab 0.934 &         0.934 &         0.933 &         0.933 \\
   & dress-sales &         0.592 &  \bftab 0.621 &         0.614 &         0.608 \\
   & exasens &         0.733 &         0.752 &         0.757 &  \bftab 0.757 \\
   & hcc &         0.734 &  \bftab 0.761 &         0.745 &         0.751 \\
   & heart-disease &         0.859 &         0.863 &  \bftab 0.871 &         0.869 \\
   & hepatitis &  \bftab 0.817 &         0.804 &         0.810 &         0.798 \\
   & horse-colic &         0.769 &         0.768 &  \bftab 0.784 &         0.783 \\
   & mammographic-masses &         0.850 &         0.852 &  \bftab 0.859 &         0.856 \\
   & mi &         0.642 &         0.636 &         0.637 &  \bftab 0.646 \\
   & nomao &         0.989 &         0.992 &  \bftab 0.994 &         0.994 \\
   & primary-tumor &         0.761 &         0.755 &  \bftab 0.767 &         0.767 \\
   & secom &         0.675 &  \bftab 0.694 &         0.679 &         0.680 \\
   & soybean &         0.997 &         0.998 &         0.999 &  \bftab 0.999 \\
   & thyroid0387 &         0.885 &         0.899 &         0.918 &  \bftab 0.928 \\
\midrule
RF & adult &         0.890 &         0.890 &         0.890 &  \bftab 0.897 \\
   & agaricus-lepiota &  \bftab 1.000 &  \bftab 1.000 &  \bftab 1.000 &  \bftab 1.000 \\
   & aps-failure &         0.988 &         0.988 &         0.989 &  \bftab 0.989 \\
   & arrhythmia &         0.885 &  \bftab 0.890 &         0.887 &         0.885 \\
   & bands &         0.885 &         0.894 &  \bftab 0.896 &         0.894 \\
   & ckd &         1.000 &         0.999 &         1.000 &  \bftab 1.000 \\
   & crx &         0.930 &         0.931 &         0.931 &  \bftab 0.931 \\
   & dress-sales &         0.585 &  \bftab 0.609 &         0.606 &         0.576 \\
   & exasens &         0.734 &  \bftab 0.753 &         0.702 &         0.707 \\
   & hcc &         0.794 &         0.806 &  \bftab 0.816 &         0.813 \\
   & heart-disease &         0.857 &         0.859 &  \bftab 0.864 &         0.858 \\
   & hepatitis &         0.880 &  \bftab 0.886 &         0.886 &         0.875 \\
   & horse-colic &         0.783 &         0.786 &         0.792 &  \bftab 0.798 \\
   & mammographic-masses &         0.812 &         0.812 &         0.822 &  \bftab 0.825 \\
   & mi &         0.664 &         0.679 &         0.686 &  \bftab 0.696 \\
   & nomao &         0.987 &         0.990 &         0.994 &  \bftab 0.994 \\
   & primary-tumor &         0.749 &         0.757 &  \bftab 0.758 &         0.756 \\
   & secom &         0.709 &         0.715 &         0.709 &  \bftab 0.722 \\
   & soybean &         0.997 &         0.999 &         0.999 &  \bftab 0.999 \\
   & thyroid0387 &         0.978 &         0.987 &  \bftab 0.994 &         0.992 \\
\end{longtable}

\begin{longtable}{llp{.063\linewidth}p{.063\linewidth}p{.063\linewidth}p{.063\linewidth}}
\caption{Distance-based classifiers, optimised hyperparameter values. \textbf{Bold}: highest value per distance measure.}
\label{tab_nn_optimised_auroc}\\
\toprule
Classifier & Dataset & \multicolumn{2}{l}{Boscovich distance} & \multicolumn{2}{l}{Euclidean distance} \\
      &  &     MMI-I &         PE &     MMI-I &         PE \\
\midrule
\endfirsthead
\caption[]{Distance-based classifiers, optimised hyperparameter values. \textbf{Bold}: highest value per distance measure.} \\
\toprule
Classifier & Dataset & \multicolumn{2}{l}{Boscovich distance} & \multicolumn{2}{l}{Euclidean distance} \\
      &  &     MMI-I &         PE &     MMI-I &         PE \\
\midrule
\endhead
\midrule
\multicolumn{6}{r}{{Continued on next page}} \\
\midrule
\endfoot

\bottomrule
\endlastfoot
NN & adult &         0.886 &  \bftab 0.887 &         0.886 &  \bftab 0.887 \\
      & agaricus-lepiota &  \bftab 1.000 &  \bftab 1.000 &  \bftab 1.000 &  \bftab 1.000 \\
      & aps-failure &  \bftab 0.969 &         0.969 &         0.966 &  \bftab 0.966 \\
      & arrhythmia &  \bftab 0.797 &         0.796 &         0.785 &  \bftab 0.786 \\
      & bands &         0.798 &  \bftab 0.814 &         0.794 &  \bftab 0.803 \\
      & ckd &         0.996 &  \bftab 0.999 &         0.994 &  \bftab 0.996 \\
      & crx &  \bftab 0.913 &         0.913 &         0.909 &  \bftab 0.909 \\
      & dress-sales &  \bftab 0.612 &         0.607 &  \bftab 0.614 &         0.607 \\
      & exasens &         0.728 &  \bftab 0.729 &  \bftab 0.735 &         0.729 \\
      & hcc &         0.758 &  \bftab 0.761 &         0.717 &  \bftab 0.744 \\
      & heart-disease &  \bftab 0.859 &         0.858 &         0.847 &  \bftab 0.847 \\
      & hepatitis &  \bftab 0.860 &         0.858 &  \bftab 0.862 &         0.861 \\
      & horse-colic &         0.762 &  \bftab 0.771 &         0.757 &  \bftab 0.765 \\
      & mammographic-masses &         0.838 &  \bftab 0.839 &  \bftab 0.841 &         0.839 \\
      & mi &         0.608 &  \bftab 0.626 &         0.610 &  \bftab 0.632 \\
      & nomao &         0.985 &  \bftab 0.988 &         0.983 &  \bftab 0.986 \\
      & primary-tumor &  \bftab 0.762 &         0.736 &  \bftab 0.764 &         0.749 \\
      & secom &         0.637 &  \bftab 0.688 &  \bftab 0.598 &         0.594 \\
      & soybean &         0.995 &  \bftab 0.996 &         0.994 &  \bftab 0.995 \\
      & thyroid0387 &         0.913 &  \bftab 0.915 &         0.912 &  \bftab 0.913 \\
\midrule
NN-D & adult &         0.881 &  \bftab 0.885 &         0.874 &  \bftab 0.878 \\
      & agaricus-lepiota &  \bftab 1.000 &  \bftab 1.000 &  \bftab 1.000 &  \bftab 1.000 \\
      & aps-failure &  \bftab 0.971 &         0.971 &         0.968 &  \bftab 0.969 \\
      & arrhythmia &         0.814 &  \bftab 0.815 &  \bftab 0.807 &         0.800 \\
      & bands &         0.825 &  \bftab 0.856 &         0.815 &  \bftab 0.828 \\
      & ckd &         0.997 &  \bftab 0.999 &         0.996 &  \bftab 0.998 \\
      & crx &         0.913 &  \bftab 0.917 &         0.911 &  \bftab 0.915 \\
      & dress-sales &  \bftab 0.604 &         0.597 &  \bftab 0.601 &         0.597 \\
      & exasens &  \bftab 0.731 &         0.729 &  \bftab 0.727 &         0.726 \\
      & hcc &  \bftab 0.776 &         0.764 &         0.754 &  \bftab 0.771 \\
      & heart-disease &  \bftab 0.863 &         0.863 &         0.852 &  \bftab 0.853 \\
      & hepatitis &  \bftab 0.864 &         0.860 &  \bftab 0.862 &         0.856 \\
      & horse-colic &         0.771 &  \bftab 0.771 &         0.773 &  \bftab 0.777 \\
      & mammographic-masses &         0.828 &  \bftab 0.830 &  \bftab 0.825 &         0.824 \\
      & mi &         0.610 &  \bftab 0.627 &         0.611 &  \bftab 0.633 \\
      & nomao &         0.988 &  \bftab 0.990 &         0.987 &  \bftab 0.989 \\
      & primary-tumor &  \bftab 0.779 &         0.745 &  \bftab 0.779 &         0.761 \\
      & secom &         0.662 &  \bftab 0.714 &         0.589 &  \bftab 0.621 \\
      & soybean &         0.998 &  \bftab 0.998 &         0.997 &  \bftab 0.998 \\
      & thyroid0387 &         0.914 &  \bftab 0.921 &         0.911 &  \bftab 0.913 \\
\midrule
FRNN & adult &         0.875 &  \bftab 0.881 &         0.864 &  \bftab 0.869 \\
      & agaricus-lepiota &  \bftab 1.000 &  \bftab 1.000 &  \bftab 1.000 &  \bftab 1.000 \\
      & aps-failure &         0.944 &  \bftab 0.952 &         0.962 &  \bftab 0.968 \\
      & arrhythmia &  \bftab 0.879 &         0.875 &         0.858 &  \bftab 0.868 \\
      & bands &         0.833 &  \bftab 0.870 &         0.816 &  \bftab 0.832 \\
      & ckd &         0.998 &  \bftab 1.000 &         0.997 &  \bftab 0.999 \\
      & crx &         0.917 &  \bftab 0.920 &         0.919 &  \bftab 0.919 \\
      & dress-sales &  \bftab 0.613 &         0.601 &  \bftab 0.608 &         0.596 \\
      & exasens &         0.741 &  \bftab 0.741 &         0.739 &  \bftab 0.744 \\
      & hcc &         0.781 &  \bftab 0.781 &         0.771 &  \bftab 0.775 \\
      & heart-disease &         0.856 &  \bftab 0.862 &         0.846 &  \bftab 0.852 \\
      & hepatitis &         0.876 &  \bftab 0.881 &         0.871 &  \bftab 0.878 \\
      & horse-colic &         0.757 &  \bftab 0.768 &         0.766 &  \bftab 0.777 \\
      & mammographic-masses &         0.845 &  \bftab 0.853 &         0.843 &  \bftab 0.847 \\
      & mi &         0.648 &  \bftab 0.665 &         0.648 &  \bftab 0.653 \\
      & nomao &         0.987 &  \bftab 0.991 &         0.987 &  \bftab 0.990 \\
      & primary-tumor &         0.777 &  \bftab 0.778 &  \bftab 0.784 &         0.780 \\
      & secom &         0.619 &  \bftab 0.667 &         0.572 &  \bftab 0.585 \\
      & soybean &  \bftab 0.997 &         0.997 &         0.996 &  \bftab 0.997 \\
      & thyroid0387 &         0.898 &  \bftab 0.914 &         0.894 &  \bftab 0.905 \\
\midrule
SVM-G & adult &               &               &         0.900 &  \bftab 0.906 \\
      & agaricus-lepiota &               &               &  \bftab 1.000 &  \bftab 1.000 \\
      & aps-failure &               &               &         0.970 &  \bftab 0.972 \\
      & arrhythmia &               &               &         0.877 &  \bftab 0.882 \\
      & bands &               &               &         0.846 &  \bftab 0.872 \\
      & ckd &               &               &  \bftab 1.000 &  \bftab 1.000 \\
      & crx &               &               &         0.916 &  \bftab 0.919 \\
      & dress-sales &               &               &  \bftab 0.634 &         0.621 \\
      & exasens &               &               &         0.779 &  \bftab 0.781 \\
      & hcc &               &               &  \bftab 0.793 &         0.778 \\
      & heart-disease &               &               &         0.864 &  \bftab 0.873 \\
      & hepatitis &               &               &         0.834 &  \bftab 0.841 \\
      & horse-colic &               &               &         0.787 &  \bftab 0.788 \\
      & mammographic-masses &               &               &         0.854 &  \bftab 0.855 \\
      & mi &               &               &         0.660 &  \bftab 0.669 \\
      & nomao &               &               &         0.992 &  \bftab 0.992 \\
      & primary-tumor &               &               &  \bftab 0.792 &         0.786 \\
      & secom &               &               &         0.672 &  \bftab 0.680 \\
      & soybean &               &               &  \bftab 0.999 &         0.999 \\
      & thyroid0387 &               &               &         0.955 &  \bftab 0.961 \\
\end{longtable}

\begin{longtable}{llp{.063\linewidth}p{.063\linewidth}}
\caption{Decision tree classifiers, optimised hyperparameter values. \textbf{Bold}: highest value.}
\label{tab_dt_optimised_auroc}\\
\toprule
Classifier & Dataset &     MMI-I &         PE \\
\midrule
\endfirsthead
\caption[]{Decision tree classifiers, optimised hyperparameter values. \textbf{Bold}: highest value.} \\
\toprule
Classifier & Dataset &     MMI-I &         PE \\
\midrule
\endhead
\midrule
\multicolumn{4}{r}{{Continued on next page}} \\
\midrule
\endfoot

\bottomrule
\endlastfoot
ABT & adult &         0.915 &  \bftab 0.915 \\
   & agaricus-lepiota &  \bftab 1.000 &  \bftab 1.000 \\
   & aps-failure &         0.986 &  \bftab 0.987 \\
   & arrhythmia &  \bftab 0.741 &         0.724 \\
   & bands &  \bftab 0.812 &         0.807 \\
   & ckd &         1.000 &  \bftab 1.000 \\
   & crx &  \bftab 0.909 &         0.909 \\
   & dress-sales &  \bftab 0.576 &         0.572 \\
   & exasens &  \bftab 0.721 &         0.716 \\
   & hcc &         0.722 &  \bftab 0.735 \\
   & heart-disease &  \bftab 0.863 &         0.862 \\
   & hepatitis &  \bftab 0.803 &         0.790 \\
   & horse-colic &         0.753 &  \bftab 0.767 \\
   & mammographic-masses &         0.855 &  \bftab 0.856 \\
   & mi &  \bftab 0.600 &         0.586 \\
   & nomao &         0.987 &  \bftab 0.987 \\
   & primary-tumor &  \bftab 0.769 &         0.757 \\
   & secom &  \bftab 0.661 &         0.657 \\
   & soybean &         0.972 &  \bftab 0.990 \\
   & thyroid0387 &         0.878 &  \bftab 0.888 \\
\midrule
CART & adult &         0.881 &  \bftab 0.882 \\
   & agaricus-lepiota &         0.996 &  \bftab 0.999 \\
   & aps-failure &         0.974 &  \bftab 0.975 \\
   & arrhythmia &         0.723 &  \bftab 0.727 \\
   & bands &         0.750 &  \bftab 0.752 \\
   & ckd &  \bftab 0.986 &         0.981 \\
   & crx &  \bftab 0.909 &         0.902 \\
   & dress-sales &  \bftab 0.596 &         0.592 \\
   & exasens &         0.733 &  \bftab 0.734 \\
   & hcc &         0.622 &  \bftab 0.627 \\
   & heart-disease &         0.798 &  \bftab 0.800 \\
   & hepatitis &         0.755 &  \bftab 0.767 \\
   & horse-colic &         0.709 &  \bftab 0.718 \\
   & mammographic-masses &  \bftab 0.838 &         0.835 \\
   & mi &  \bftab 0.632 &         0.631 \\
   & nomao &  \bftab 0.965 &         0.964 \\
   & primary-tumor &  \bftab 0.713 &         0.712 \\
   & secom &         0.642 &  \bftab 0.658 \\
   & soybean &         0.963 &  \bftab 0.964 \\
   & thyroid0387 &         0.918 &  \bftab 0.920 \\
\midrule
ERT & adult &         0.893 &  \bftab 0.898 \\
   & agaricus-lepiota &  \bftab 1.000 &         1.000 \\
   & aps-failure &  \bftab 0.985 &         0.985 \\
   & arrhythmia &  \bftab 0.872 &         0.867 \\
   & bands &         0.848 &  \bftab 0.857 \\
   & ckd &         1.000 &  \bftab 1.000 \\
   & crx &         0.919 &  \bftab 0.922 \\
   & dress-sales &  \bftab 0.631 &         0.618 \\
   & exasens &         0.765 &  \bftab 0.765 \\
   & hcc &         0.772 &  \bftab 0.774 \\
   & heart-disease &  \bftab 0.860 &         0.859 \\
   & hepatitis &         0.864 &  \bftab 0.868 \\
   & horse-colic &  \bftab 0.793 &         0.790 \\
   & mammographic-masses &         0.858 &  \bftab 0.859 \\
   & mi &         0.680 &  \bftab 0.683 \\
   & nomao &         0.980 &  \bftab 0.981 \\
   & primary-tumor &  \bftab 0.791 &         0.787 \\
   & secom &  \bftab 0.729 &         0.728 \\
   & soybean &         0.998 &  \bftab 0.998 \\
   & thyroid0387 &         0.967 &  \bftab 0.970 \\
\midrule
GBM & adult &         0.918 &  \bftab 0.919 \\
   & agaricus-lepiota &  \bftab 1.000 &         1.000 \\
   & aps-failure &         0.987 &  \bftab 0.987 \\
   & arrhythmia &  \bftab 0.905 &         0.904 \\
   & bands &         0.862 &  \bftab 0.862 \\
   & ckd &         1.000 &  \bftab 1.000 \\
   & crx &         0.937 &  \bftab 0.937 \\
   & dress-sales &  \bftab 0.609 &         0.602 \\
   & exasens &         0.783 &  \bftab 0.784 \\
   & hcc &  \bftab 0.794 &         0.790 \\
   & heart-disease &         0.874 &  \bftab 0.874 \\
   & hepatitis &         0.863 &  \bftab 0.873 \\
   & horse-colic &         0.788 &  \bftab 0.789 \\
   & mammographic-masses &  \bftab 0.866 &         0.862 \\
   & mi &         0.704 &  \bftab 0.709 \\
   & nomao &         0.990 &  \bftab 0.990 \\
   & primary-tumor &  \bftab 0.794 &         0.792 \\
   & secom &  \bftab 0.718 &         0.713 \\
   & soybean &         0.999 &  \bftab 0.999 \\
   & thyroid0387 &  \bftab 0.955 &         0.951 \\
\midrule
RF & adult &         0.904 &  \bftab 0.907 \\
   & agaricus-lepiota &         1.000 &  \bftab 1.000 \\
   & aps-failure &  \bftab 0.985 &         0.985 \\
   & arrhythmia &  \bftab 0.864 &         0.857 \\
   & bands &  \bftab 0.860 &         0.860 \\
   & ckd &         0.999 &  \bftab 1.000 \\
   & crx &         0.930 &  \bftab 0.931 \\
   & dress-sales &  \bftab 0.640 &         0.627 \\
   & exasens &  \bftab 0.763 &         0.757 \\
   & hcc &  \bftab 0.818 &         0.804 \\
   & heart-disease &  \bftab 0.862 &         0.859 \\
   & hepatitis &  \bftab 0.872 &         0.872 \\
   & horse-colic &         0.781 &  \bftab 0.785 \\
   & mammographic-masses &         0.860 &  \bftab 0.862 \\
   & mi &         0.669 &  \bftab 0.678 \\
   & nomao &  \bftab 0.979 &         0.979 \\
   & primary-tumor &         0.787 &  \bftab 0.791 \\
   & secom &         0.709 &  \bftab 0.712 \\
   & soybean &         0.998 &  \bftab 0.998 \\
   & thyroid0387 &  \bftab 0.975 &         0.975 \\
\end{longtable}

\end{document}